\documentclass[letterpaper,journal]{IEEEtran}
\usepackage{amsmath,amsfonts}
\usepackage{algorithmic}
\usepackage{array}
\usepackage{textcomp}
\usepackage{stfloats}
\usepackage{url}
\usepackage{verbatim}
\usepackage{enumerate}
\usepackage{graphicx}
\usepackage{cite}
\usepackage{epstopdf}
\usepackage[caption=false,font=footnotesize,labelfont=rm,textfont=rm]{subfig}
\usepackage{ragged2e}
\usepackage{booktabs,makecell,multirow,tabularx}
\usepackage[linesnumbered,ruled]{algorithm2e}
\usepackage{csquotes}
\usepackage[hidelinks]{hyperref}
\usepackage{balance}
\SetKwRepeat{Do}{do}{while}%
\def\BibTeX{{\rm B\kern-.05em{\sc i\kern-.025em b}\kern-.08em
    T\kern-.1667em\lower.7ex\hbox{E}\kern-.125emX}}
\begin{document}
\title{Automatic programming via large language models with population self-evolution for dynamic fuzzy job shop scheduling problem}
\author{Jin Huang, Qihao Liu, Xinyu Li,~\IEEEmembership{Member,~IEEE}, Liang Gao,~\IEEEmembership{Senior Member,~IEEE}, and Yue Teng%
\thanks{This manuscript is the authors' accepted version of an article accepted for publication in \textit{IEEE Transactions on Fuzzy Systems}; the final version of record may differ. \copyright{} 2026 IEEE. Personal use of this material is permitted. Permission from IEEE must be obtained for all other uses, in any current or future media, including reprinting/republishing this material for advertising or promotional purposes, creating new collective works, for resale or redistribution to servers or lists, or reuse of any copyrighted component of this work in other works.}
\thanks{All authors are with the State Key Laboratory of Intelligent Manufacturing Equipment and Technology, Huazhong University of Science and Technology, Wuhan 430074, China. (Corresponding author: Xinyu Li.)}
}

\markboth{Accepted Manuscript}%
{Huang \MakeLowercase{\textit{et al.}}: Automatic Programming via Large Language Models for DFJSSP}

\maketitle

\begin{abstract}
Heuristic dispatching rules (HDRs) are widely used for solving the dynamic fuzzy job shop scheduling problem (DFJSSP). However, their performance is highly sensitive to specific scenarios and often necessitates expert customization.  To overcome this, automated design methods like genetic programming (GP) and gene expression programming (GEP) have been proposed. Despite their success, these methods face challenges, such as high randomness in the search process. Recently, the combination of large language models (LLMs) with evolutionary algorithms has opened new possibilities for prompt engineering and automated algorithm design. To improve the ability of LLMs in automatic HDR design, this paper introduces a novel population self-evolutionary (SeEvo) framework, which draws inspiration from the self-reflective design strategies employed by human experts. Notably, this framework employs a novel teacher-student learning mechanism, allowing the LLM (student) to generate robust HDRs. Guided by a teacher model with complete knowledge of actual processing times, the student learns to infer fuzzy uncertainties from historical deviations, enabling it to effectively anticipate and adapt to fuzzy impacts.  Experimental results demonstrate that SeEvo significantly outperforms GP, GEP, deep reinforcement learning (DRL) methods, and more than ten commonly used HDRs from the literature, particularly in previously unseen and dynamic scenarios.
\end{abstract}

\section{Introduction}
The core challenge in production scheduling is the efficient allocation of limited resources to complete tasks in a timely manner while optimizing performance metrics~\cite{shadyFeatu2023}. A classic formulation of this challenge is the job shop scheduling problem (JSSP), which is an NP-hard optimization problem. For static JSSP, where all parameters like processing times and job orders are predetermined, traditional exact methods such as dynamic programming and branch-and-bound are computationally infeasible for large-scale instances~\cite{zhangMathe2019, huang2025novel, wang2018nsga}. Consequently, researchers have largely shifted towards metaheuristic algorithms, which can yield near-optimal solutions within a reasonable timeframe~\cite{yao2024dqn, zhang2025tackling}.

However, real-world manufacturing environments are rarely static and often subject to dynamic uncertainties, such as unexpected machine breakdowns, the arrival of new orders, or fuzzy processing time. This has led to growing interest in the dynamic fuzzy JSSP (DFJSSP) \cite{gao2020solving, wang2022solving, wang2023data, wu2024deep}. In such dynamic and uncertain production environments, heuristic dispatching rules (HDRs) are often preferred for their low computational complexity and rapid responsiveness \cite{holthausEffic1997}. The primary challenge lies in designing effective HDRs, a process that has traditionally relied on expert knowledge and extensive trial-and-error. To overcome this limitation, automated approaches for discovering dispatching rules have been developed. Among these, evolutionary frameworks like genetic programming (GP) \cite{liRealt2022, wang2023data} and gene expression programming (GEP) \cite{zhangMathe2019, nieReact2013} have demonstrated potential in automating the design and optimization of HDRs.

While GP and GEP have achieved success in various scheduling scenarios, they rely primarily on stochastic search operators (e.g., random mutation and crossover) for exploration. In recent years, deep reinforcement learning (DRL) has emerged as a complementary paradigm, demonstrating strong capabilities in learning scheduling policies through environment interaction~\cite{liMulti2025,liu2022deep}. DRL methods excel particularly in fast-paced production environments requiring real-time decisions, with inference times typically under one second after training~\cite{chendeep2023, liu2022deep}. These characteristics make DRL well-suited for applications such as electronics assembly.

Despite the success of these dynamic scheduling methods, they rely on two fundamentally different exploration mechanisms: stochastic search (for GP/GEP) and environment-driven policy learning (for DRL).  The recent advent of large language models (LLMs) presents new opportunities to guide evolutionary search through semantic reasoning  \cite{romera-paredesMathe2024}. By combining prompt engineering with iterative feedback mechanisms, LLMs can generate highly adaptive and domain-specific HDRs. Leveraging extensive training data, these models capture deep domain knowledge and structural patterns \cite{dat2025hsevo, yeReEvo2024}, enabling them to produce high-quality solutions for test cases within short computation times. Research has already demonstrated the potential of LLMs for automatic algorithm design in problems like the online bin packing problem (BPP) \cite{romera-paredesMathe2024}. However, their application to more complex scheduling problems, such as DFJSSP, remains an area of exploration.

Beyond the general challenge of automated HDR design, DFJSSP presents a specific difficulty: actual processing times are often fuzzy, creating significant information gaps for decision-makers. While many dynamic scheduling methods acknowledge this uncertainty, they often address it by treating deviations as dynamic events that trigger reactive adjustments \cite{wu2024deep, zhang2023deep}, rather than proactively designing schedules that are robust to such variations. We reframe the fuzzy scheduling problem as a decision-making task under incomplete information, aiming to develop a scheduling model that can infer and anticipate the effects of uncertainty.

Therefore, this paper proposes a novel LLM-based evolutionary framework for automatic HDRs design in DFJSSP, introducing an innovative population self-evolutionary (SeEvo) method. In this framework, LLMs act as hyper-heuristic generators and employ self-evolution within the population to automatically design and optimize HDRs. Specifically, the contributions of this paper are as follows:

\begin{itemize}
    \item[1)] \textbf{Novel LLM-Based Evolutionary Framework for DFJSSP:} This paper introduces a novel LLM-based evolutionary framework that addresses the automatic design of algorithms for DFJSSP. The framework's core innovation is a training paradigm that teaches the LLM to generate robust HDRs by learning from the gap between planned and actual processing times, thereby enhancing resilience to stochastic uncertainty.\par
    \item[2)] \textbf{Population Self-Evolution Strategy:} A novel population self-evolution strategy is proposed within the LLM-based framework, significantly improving the exploration and exploitation capabilities of the generated heuristics. This strategy allows for continuous refinement of HDRs based on real-time feedback during the scheduling process, enhancing scheduling efficiency in dynamic environments. \par
    \item[3)]  \textbf{Comprehensive Evaluation and Superiority of the SeEvo Method:} The proposed SeEvo method's effectiveness and superiority are demonstrated through extensive comparisons with commonly used HDRs, GP, GEP, and DRL methods. Experimental results demonstrate the SeEvo method's superior generalization across unseen and dynamic DFJSSP, outperforming other dynamic scheduling methods.\par
\end{itemize}\par

The remainder of this paper is organized as follows. Section II  reviews the pertinent literature. Section III details the language-heuristic-based DFJSSP framework, followed by a detailed explanation of the population self-evolution method in Section IV. Section V presents the experimental setup and discusses the performance evaluation. Finally, Section VI concludes the paper and offers directions for future work.
\section{Background}
\subsection{Related Works of Fuzzy Job Shop Scheduling Problem}
Over the past several decades, many methods have been proposed to solve the JSSP. Exact methods, such as dynamic programming and branch-and-bound algorithms, can guarantee optimal solutions but are often limited to smaller-scale problems due to their high computational complexity \cite{gromicho2012solving, huang2026efficient}. In contrast, metaheuristic methods such as genetic algorithms, tabu search, and simulated annealing have been widely adopted to provide efficient solutions for larger-scale problems \cite{xienewn2023}. For the FJSSP, an effective differential evolution algorithm with an improved selection mechanism was proposed, showing promising results. Moreover, researchers have expanded their focus to multi-objective FJSSP. For instance, hybrid local simulated annealing algorithms combined with NSGA-II \cite{wang2018nsga} and hybrid particle swarm optimization with genetic algorithms \cite{gao2020solving} have shown effective results. A self-learning discrete artificial bee colony algorithm was also introduced to address fuzzy welding seam scheduling \cite{yu2024self}. Despite their success in static optimization, these methods face significant challenges in highly dynamic environments, such as frequent order arrivals and machine breakdowns, making it difficult to achieve high-quality solutions within acceptable computational times.

HDRs have gained popularity due to their simplicity, efficiency, and responsiveness to real-time changes in the scheduling environment \cite{durasevicsurve2018,holthausEffic1997}. Although widely used, their effectiveness is highly dependent on specific scenarios and often requires extensive expert knowledge for customization. To address these limitations, hyper-heuristic methods have emerged, which automatically evolve or select heuristic methods, offering more flexibility and adaptability to scheduling problems. In particular, GP \cite{liRealt2022,wang2023data} and GEP \cite{zhangMathe2019,nieReact2013} have shown promise in DJSSP. These methods enable automatic learning of scheduling heuristics without  domain-specific prior knowledge, making them suitable for DJSSP \cite{shadyFeatu2023}. However, a major challenge of these hyper-heuristics is the large feature space of DFJSSP, which can exponentially increase the search space, hindering exploration efficiency. Furthermore, heuristics generated may not generalize well to unseen DFJSSP instances, making it difficult to obtain high-quality dynamic scheduling solutions \cite{liu2022deep}.

In recent years, DRL methods have gained significant attention for workshop scheduling \cite{chendeep2023, liu2022deep, liuDynam2024}. Current DRL approaches can be categorized into rule-selection-based DRL, where agents select from predefined rules \cite{liMulti2025}, and end-to-end DRL, which directly maps states to job-machine assignments via graph neural networks \cite{chendeep2023, liuDynam2024}. Representative end-to-end methods employ heterogeneous graph neural networks (e.g., meta-path-based architectures \cite{wan2024flexible}, dual-shop frameworks \cite{bao2025end}, structure-aware encoders \cite{moon2024learning}) and attention mechanisms \cite{wang2025attention}. For DFJSSP, studies have focused on enabling agents to maximize long-term rewards through environment interaction \cite{wu2024deep, zhang2023deep}, with uncertain processing times addressed through various modeling approaches, including uniform distributions, asymmetric triangular intervals \cite{li2020improved}, and LR fuzzy numbers \cite{yu2024self}. These scheduling policies exhibit strong size-invariant properties, allowing generalization to unseen instances. Some researchers have also incorporated GP-based action spaces into DRL frameworks, combining the strengths of both paradigms \cite{liRealt2022}. While end-to-end DRL methods share our goal of direct job selection, they differ fundamentally in their mechanisms. DRL learns implicit neural network policies through extensive environment interaction and requires manual design of reward functions, action spaces, and state representations. In contrast, our language-heuristic approach leverages LLMs' pre-trained knowledge to generate interpretable heuristic code through  an evolutionary search guided by natural language.

\subsection{Automatic Algorithm Design with LLMs}
The field of automatic algorithm design aims to reduce the human effort required to develop high-performance algorithms. A primary approach within this field is the hyper-heuristic, which searches the space of heuristics rather than the solution space, with the objective of automatically selecting or generating effective algorithms. Traditional generative hyper-heuristics, such as GP \cite{liRealt2022,wang2023data} and GEP \cite{zhangMathe2019,nieReact2013}, construct new algorithms by evolving them from a predefined set of components. However, the rise of  LLMs has introduced powerful new opportunities. LLMs can automatically design the internal logic of a heuristic, often requiring only the function signature (name, inputs, and outputs), thus moving beyond the constraints of a fixed component set. This marks a conceptual leap, transforming the heuristic design process from combinatorial selection to creative generation \cite{romera-paredesMathe2024}.\par

This LLM-driven approach builds upon the demonstrated capabilities of large language models in analogous tasks, including code generation \cite{chenTeach2023} and solving algorithmic competition challenges \cite{shinnRefle2023}. The integration of LLMs with evolutionary algorithms has proven effective for various tasks, such as prompt optimization \cite{guoConne2023} and algorithm self-improvement \cite{liuAlgor2023}, and has shown particular promise in addressing combinatorial optimization problems, including the traveling salesman problem (TSP) \cite{liuexamp2024, liuAlgor2023} and the online BPP \cite{romera-paredesMathe2024}. Pioneering work in this domain includes two notable studies that were among the first to apply LLMs to the automatic design of algorithms for combinatorial optimization. Liu et al. \cite{liuAlgor2023} proposed an algorithmic self-evolution framework that leverages LLMs to automatically evolve TSP heuristics using the GPT-4o model. Similarly, Romera-Paredes et al. \cite{romera-paredesMathe2024} combined a fine-tuned PaLM2 (340B) model with an evolutionary framework and introduced a multi-island collaborative program library update mechanism to address various combinatorial optimization problems, achieving optimal results for the online BPP in their work published in \textit{Nature}. Furthermore, recent research in \textit{Nature Machine Intelligence} \cite{mischler2024contextual} demonstrates that LLMs exhibit increasingly brain-like processing capabilities, further providing additional theoretical support for their application in optimization tasks through evolutionary algorithms.

Despite the promising application of LLMs in heuristic algorithm development for TSP and BPP, their use in more complex problems, such as DFJSSP, remains limited. The ReEvo framework \cite{yeReEvo2024}, an advanced language-heuristic approach, has demonstrated success in TSP by simulating human expert reflection processes. However, the complexity of DFJSSP far exceeds that of TSP, as it involves multiple jobs across different machines, as well as dynamic events like fuzzy processing times, dynamic order arrivals, and machine breakdowns, making it a significantly challenging problem. Although the ReEvo framework has performed well in TSP, primarily due to its short-term reflection (comparative learning of individual differences) and long-term reflection (summarizing short-term reflections), it has not fully explored the potential for population self-evolution. Moreover, the framework was primarily designed for static problems and does not distinguish between training and test sets.

Given the dynamic nature of DFJSSP, with frequent changes in processing scenarios and dynamic events, this paper aims to explore a novel language-heuristic framework that fully utilizes the capabilities of LLMs to provide more efficient and adaptable HDRs. However, recent LLM-based hyper-heuristic approaches face a critical limitation for practical deployment: they rely on lengthy iterative evolution processes that typically require tens of minutes to generate high-performance rules, making them impractical for dynamic environments where workshop conditions can change within minutes. To address this timeliness bottleneck, our framework introduces a knowledge distillation perspective that enables rapid response capabilities essential for real-world dynamic scheduling scenarios.\par

\section{Framework of Language-Heuristic-Based DFJSSP}
\begin{figure}[htbp]
	\centering
	\includegraphics[width=\linewidth]{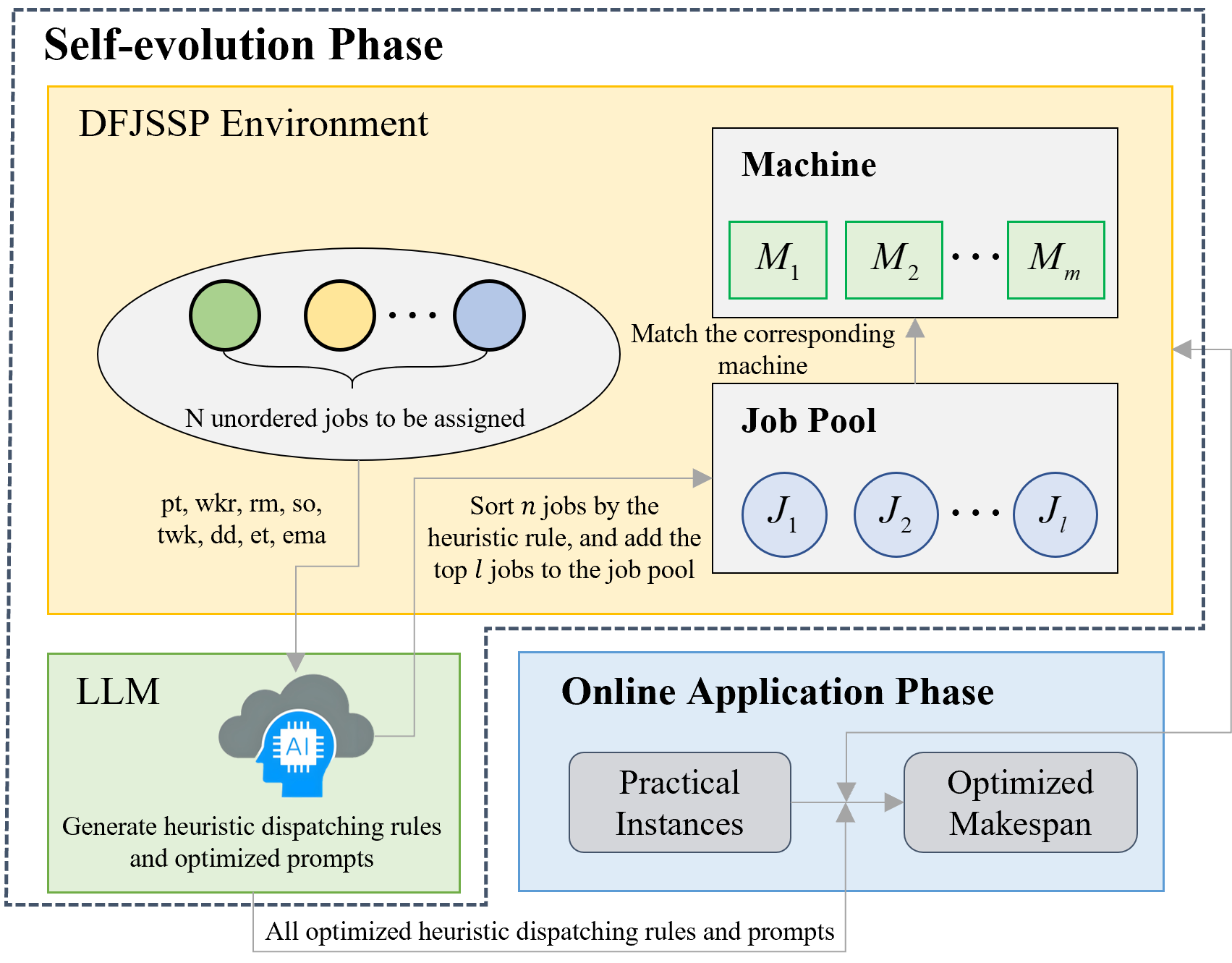}			
	\caption{Language-heuristic-based DJSSP framework.}
	\label{fig1}
\end{figure}

The framework for the language-heuristic-based dynamic DFJSSP, as shown in Fig.\ref{fig1}, consists of two stages: the self-evolution stage and the online application stage. To address the DFJSSP with randomly arriving orders, fuzzy processing times, and machine breakdowns, a job shop simulation environment is designed to handle these uncertainties and dynamic conditions. Additionally, a language-heuristic-based SeEvo method is developed to evolve HDRs.\par
As illustrated in Fig.\ref{fig1}, during the self-evolution stage, the DFJSSP environment consists of a job pool and a set of machines. The statistical data of incoming jobs is initially input into the LLM, including the processing time of the current operation for each job \((pt)\), the remaining work time \((wkr)\), the remaining processing time excluding the current operation \((rm)\), the processing time of subsequent operations \((so)\), the total processing time \((twk)\), the exponential moving average \((ema)\), the due date of each job \((dd)\), and the estimated tardiness \((et)\) \cite{li2025real}. Based on the statistical features, the LLM generates the HDR, which is then used in the DFJSSP environment for job sequencing. Only the top-ranked jobs are available for the scheduling process. Once a job's operation is completed, the machine becomes available and selects a new job from the job pool. During this stage, the LLM collects substantial training data and updates HDRs to improve the scheduling solution.

A core innovation of this framework is its handling of the fuzzy deviation between planned and actual processing times. In real-world scenarios, the scheduling algorithm only knows the planned time, while the actual time remains uncertain. To bridge this gap, we introduce a learning mechanism based on the \enquote{teacher-student} paradigm. In our simulation, the environment acts as the \enquote{teacher}, with knowledge of the actual processing time \( p'_{ij} \). The \enquote{student} only knows the planned processing time \( p_{ij} \) and other observable states. The key mechanism enabling the \enquote{student} to anticipate fuzzy uncertainties is the $ema$ feature. The relative deviation, defined as $\frac{p'_{ij}}{p_{ij}} - 1$, is tracked using the exponential fuzzy time smoothing formula, which adjusts statistical metrics in real-time:
\begin{equation}
\varphi_{ij}=\kappa_{ij} \delta_{ij}+\left(1-\kappa_{ij}\right) \varphi_{i-1j}
\end{equation}
where $\kappa_{ij}$ is the fuzzy adaptive adjustment coefficient (set to 0.2 in this paper), and $\delta_{ij}$ represents the latest relative deviation. The updated smoothing factor $\varphi_{ij}$ is then mapped back to time-related metrics such as \(pt\) and \(wkr\), creating the $ema$ feature. By injecting $ema$ into the LLM's prompt, we enable the HDR to leverage the historical deviation trend. This process allows the LLM to correlate this trend with scheduling performance, enabling HDRs to implicitly account for future time fluctuations, thereby enhancing the robustness and foresight of the proposed dynamic scheduling method.

The self-evolution stage serves as a training phase, where numerous cases are processed to accumulate experience and refine HDRs. In contrast, the online application phase utilizes the best HDRs generated during the self-evolution phase (trained on 20 selected cases in our experiments) and applies them to real-world scenarios. After a single iteration of the self-evolution framework, high-quality HDRs are produced for practical use. For instance, when tested with a job set of 100 jobs and 20 machines, the framework provides a high-quality solution within 40 seconds.

To further improve the generation of high-quality HDRs in the self-evolution stage, a novel SeEvo method is introduced. This method enhances the population’s exploration capabilities through individual co-evolution reflection, individual self-evolution reflection, and collective evolution reflection strategies. These mechanisms are explained in greater detail in the following sections.

\section{Language-Heuristic-Based Population Self-Evolution Method}
\begin{figure*}[htbp]
	\centering
	\includegraphics[width=\linewidth]{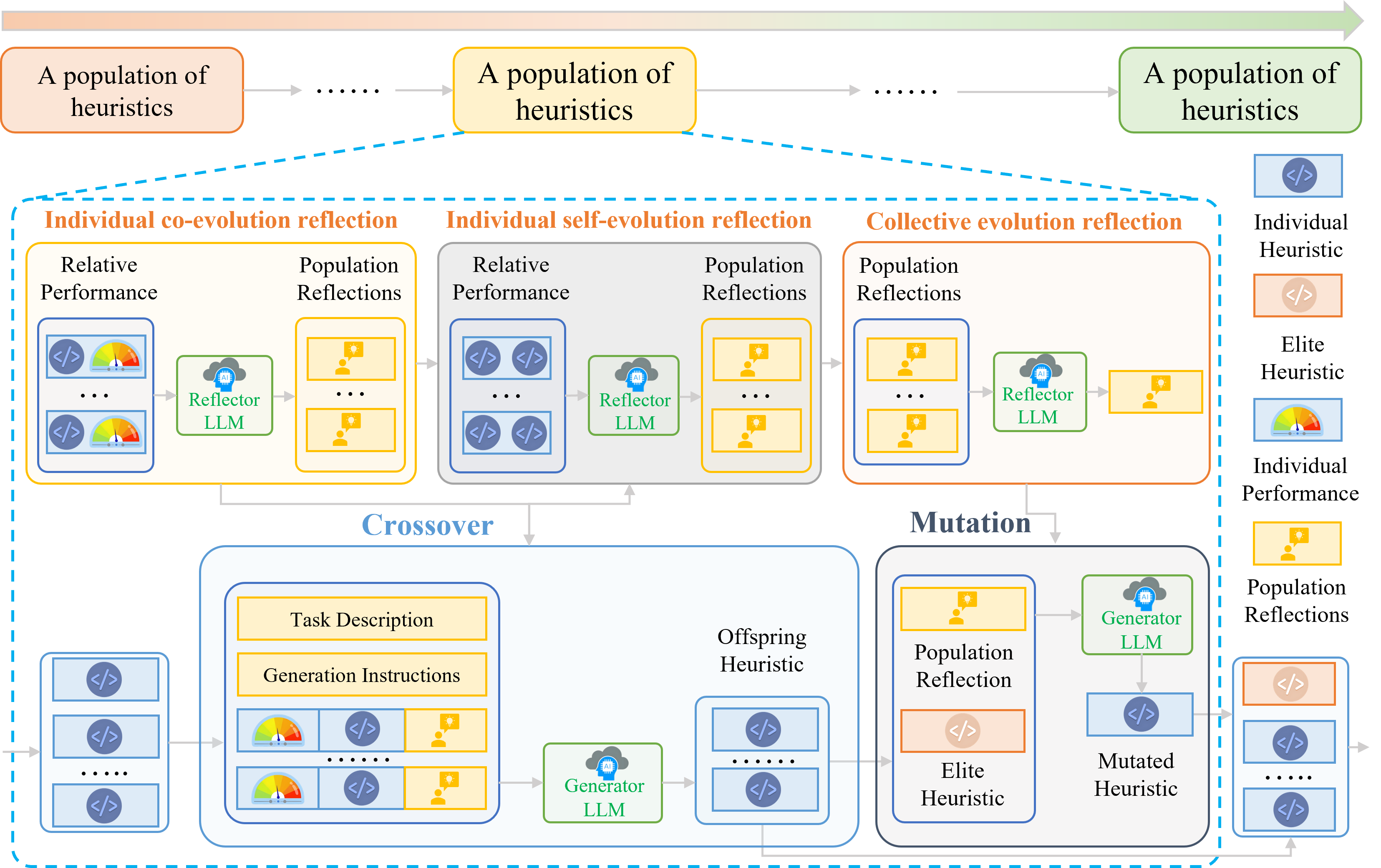}			
	\caption{The population self-evolution method.}
	\label{fig2}
\end{figure*}

The structure of the SeEvo framework, a language-driven heuristic evolution method, is outlined in Fig. \ref{fig2}. In this framework, LLMs play two key roles: generating guiding prompts for the population and creating individual heuristic code segments. Unlike traditional hyper-heuristic methods, SeEvo relies on the independent generation of heuristic code fragments, which are continuously refined through the evolutionary process. SeEvo leverages LLMs to generate guiding prompts and craft HDRs tailored to specific scheduling tasks.

SeEvo follows a multi-stage iterative procedure, with each phase contributing to the progressive refinement of the heuristic population to enhance job selection efficiency. Further details on the prompt design process, along with an overview of the stages, can be found in \textbf{Supplementary Material}, Fig.S1. This includes the corresponding design prompt for the population self-evolution method, specifically aimed at optimizing makespan. To optimize for tardiness instead, the \texttt{problem\_desc} section can be modified by replacing the optimization objective with tardiness.

The core of SeEvo comprises a set of reflection operators that replace and enhance traditional genetic operators. These operators generate and synthesize language-based knowledge, which we denote as $R$.

\subsection{Individual Encoding and Initialization}
Unlike traditional evolutionary algorithms, SeEvo employs a flexible encoding mechanism where individuals are code fragments rather than fixed-length representations. In SeEvo, individuals are HDRs designed to guide job selection in the DFJSSP, rather than directly determining the final scheduling plan. These HDRs are generated by the LLM without constraints on encoding length or predefined function sets. The only requirement is that the LLM-generated code must adhere to specific function names, input parameters, and output parameters. For instance, the \texttt{Seed\_func} represents an initial HDR, where the function is named \texttt{get\_combined\_expression\_v1}, with inputs including $pt$, $wkr$, $rm$, $so$, $twk$, $ema$, $dd$, and $et$, and the output is \texttt{combined\_expression\_data}, which is generated by combining the input parameters. The population initialization in SeEvo is facilitated by the LLM prompt generator, which uses the \texttt{seed\_func} along with detailed generation instructions. These instructions specify the function name and the associated heuristic logic. The seed heuristics, which are example heuristic codes, form the foundation for guiding the LLM in generating initial HDRs in promising search directions.

\subsection{The Reflection-Evolution Loop}
The framework operates through an iterative loop in which individuals are evaluated, reflected upon, and evolved to generate superior candidates. This process is orchestrated by three reflection mechanisms and three evolution operators.

\subsubsection{Individual Co-Evolution Reflection ($R_{\text{co}}$)}
This mechanism compares two randomly selected HDRs to analyze their performance disparities. The reflector LLM takes two parent individuals, $H_i$ and $H_j$, along with their performance metrics ($\text{Perf}_i, \text{Perf}_j$), and generates a comparative reflection $R_{\text{co}}$ that captures the underlying reasons for their performance difference:
\begin{equation}
R_{\text{co}} = \text{LLM}_{\text{Reflect}}(\text{Prompt}_{\text{co-reflect}}, H_i, H_j, \text{Perf}_i, \text{Perf}_j)
\end{equation}
where this $R_{\text{co}}$ subsequently serves as guidance for the Crossover operator.

\subsubsection{Individual Self-Evolution Reflection ($R_{\text{self}}$)}
This mechanism constitutes the core contribution of SeEvo. Following the generation of offspring through crossover, it compares each offspring $H_{\text{child}}$ with its corresponding parent $H_{\text{parent}}$ from the selected population. The performance differential $\Delta P = \text{Perf}(H_{\text{child}}) - \text{Perf}(H_{\text{parent}})$ is analyzed to generate a self-reflection:
\begin{equation}
R_{\text{self}} = \text{LLM}_{\text{Reflect}}(\text{Prompt}_{\text{self-reflect}}, H_{\text{parent}}, H_{\text{child}}, \Delta P)
\end{equation}
where $R_{\text{self}}$ functions as a failure analysis to prevent recurring errors when $\Delta P \leq 0$, while reinforcing successful strategies when $\Delta P > 0$. This reflection subsequently guides the Self-Evolution process within the Crossover operator.

\subsubsection{Collective Evolution Reflection ($R_{\text{coll}}$)}
This module serves as the framework's long-term memory by synthesizing all local reflections into accumulated evolutionary knowledge. At the conclusion of each generation $t$, all local insights form a reflection set $\mathcal{R}_{\text{gen\_t}} = \{R_{\text{co}}^1, \dots, R_{\text{co}}^M\} \cup \{R_{\text{self}}^1, \dots, R_{\text{self}}^N\}$, which is synthesized with the memory from the previous generation:
\begin{equation}
R_{\text{coll}}^{(t)} = \text{LLM}_{\text{Synthesize}}(\text{Prompt}_{\text{collective}}, R_{\text{coll}}^{(t-1)}, \mathcal{R}_{\text{gen\_t}})
\end{equation}
where the updated $R_{\text{coll}}^{(t)}$ then guides all evolution operators in the subsequent generation.

\subsubsection{Crossover}
The Crossover operator performs intelligent semantic fusion guided by both local and global reflections, operating through two distinct processes: Co-Evolution Crossover and Self-Evolution refinement.

The Co-Evolution Crossover receives parent HDRs ($H_i, H_j$), their corresponding Co-Evolution Reflection ($R_{\text{co}}$), and the collective wisdom ($R_{\text{coll}}^{(t-1)}$):
\begin{equation}
H_{\text{child}} = \text{LLM}_{\text{Crossover}}(\text{Prompt}_{\text{cross}}, H_i, H_j, R_{\text{co}}, R_{\text{coll}}^{(t-1)})
\end{equation}
where this dual-guidance mechanism enables strategy-level reasoning to coherently synthesize complementary strengths from both parents.

The Self-Evolution process then refines each Co-Evolution offspring. For each offspring $H_{\text{child}}$ and its corresponding parent $H_{\text{parent}}$, the Crossover operator utilizes the Self-Evolution Reflection ($R_{\text{self}}$) and collective memory:
\begin{equation}
\resizebox{0.98\linewidth}{!}{$
H^{\text{new}}_{\text{child}}= \text{LLM}_{\text{Crossover}}(\text{Prompt}_{\text{cross}}, H_{\text{parent}}, H_{\text{child}}, R_{\text{self}}, R_{\text{coll}}^{(t-1)})
$}
\end{equation}

Through this two-stage crossover approach, the algorithm learns from both successful and failed evolutionary attempts, progressively improving offspring quality via reflection-guided refinement.

\subsubsection{Mutation}
The Mutation operator transforms random perturbation into knowledge-driven exploration. It leverages the collective memory ($R_{\text{coll}}^{(t)}$) to guide strategic modifications of elite individuals ($H_{\text{elite}}$):
\begin{equation}
H_{\text{mutant}} = \text{LLM}_{\text{Mutate}}(\text{Prompt}_{\text{mutate}}, H_{\text{elite}}, R_{\text{coll}}^{(t)})
\end{equation}
where this approach enables intelligent modifications grounded in accumulated evolutionary knowledge.

\subsubsection{Individual Evaluation}
Each HDR undergoes rigorous evaluation with respect to the scheduling task, where optimization objectives encompass makespan  and tardiness minimization. This evaluation mechanism ensures continuous refinement of the population toward optimal solutions. Detailed design specifications are provided in Fig.~S2 of the \textbf{Supplementary Material}.

\subsection{The SeEvo Algorithm Framework}
Due to space limitations, the specific design process of Algorithm 1 is provided in Fig.S3 of the \textbf{Supplementary Material}. This pseudocode clarifies the precise loop of how reflections are generated, collected, synthesized, and then fed back into the generative operators (Crossover and Mutation) in the subsequent generation.

\section{Experimental Evaluation}
\subsection{Experimental Setup}
To validate the effectiveness of the proposed SeEvo method, experiments are conducted under both static and dynamic conditions. For the static experiments, publicly available benchmark datasets from Taillard (TA) and Demirkol (DMU) are used. For the dynamic experiments, a DFJSSP environment with randomly generated order variations, machine breakdowns, and fuzzy processing times is simulated. To comprehensively evaluate the performance of the proposed method, both makespan minimization and tardiness minimization are considered. Regarding the use of LLM, we employ three APIs: gpt-4.1-mini-2025-04-14, Qwen-Turbo, and DeepSeek-V3. It is worth mentioning that DeepSeek-R1 is excluded from comparison due to its slow response time. Additionally, the SeEvo method is compared with the GEP algorithm \cite{nieReact2013}, the GP algorithm \cite{liRealt2022}, over ten common HDRs, and end-to-end DRL methods \cite{chendeep2023,liuDynam2024,zhangLearn2020c}. Note that the GEP and GP methods in the literature are primarily for flexible JSSP, and we reference them only for the job selection, using similar statistical feature selections as in SeEvo. Lastly, an ablation study is conducted to validate the effectiveness.
\begin{table}[htbp]
  \centering
  \caption{Parameters of SeEvo}
    \begin{tabular}{cc}
    \toprule
    Parameter & Value \\
    \midrule
    LLM & \{gpt, Qwen, DeepSeek\} \\
    LLM temperature & \{0.0, 0.5, 1.0\} \\
    Population size & \{10, 15, 20\} \\
    Maximum iterations & \{10, 15, 20\} \\
    Mutation rate & \{0.5, 0.7, 0.9\} \\
    \bottomrule
    \end{tabular}%
  \label{tab:addlabe0}%
\end{table}%

In the experiments, the parameters for the SeEvo method are chosen using orthogonal experiments (Taguchi design), as shown in Table~\ref{tab:addlabe0}. Both static and dynamic training cases are randomly generated as follows:

\textbf{Static Dataset:} SeEvo’s generalization performance is first assessed using randomly generated training cases, varying in size from 20 to 100 jobs and 10 to 20 machines. The processing times for each job are randomly assigned between 50 and 100 units.

\textbf{Dynamic Dataset:} For dynamic scheduling scenarios, 20 training cases are generated, with the number of jobs in each batch randomly ranging between 20 and 50. Each batch consists of 2 to 3 sub-batches, with arrival times uniformly distributed within [1, 500] and [501, 1000]. The DFJSSP environment includes 5 to 20 machines, and job processing times are randomly assigned between 50 and 100 units. Machine breakdowns occur randomly, with 1 to 2 breakdowns per case. The breakdown durations range from 1 to 500 time units, and the associated repair times range from 10 to 50 units. To simulate real-world scenarios, gamma-distributed noise perturbation is applied to the job processing times, using a gamma distribution with a shape parameter of 1.0 and a scale parameter of 0.10, introducing uncertainty in processing times.

For training, 20 randomly generated cases are used, with one case replaced after each iteration. During testing, the crossover and mutation probabilities remain constant, but the number of generations is reduced to one.\par

Unlike DRL methods, the GP and GEP approaches are more tailored to specific cases and are more reliant on training data. To avoid an unfair comparison, 80 training cases are introduced for the self-generated algorithm, with 20 generations per round. After each round, the training cases are replaced. The population size for GP and GEP is set to 20, matching that of SeEvo, with 4 randomly selected cases per round. All methods are implemented in Python and executed on a server running Ubuntu 20.04 with an Intel(R) Xeon(R) W-3365 CPU @ 2.70GHz. \par

\subsection{Parameter Selection for the SeEvo Method}

Parameter selection is a crucial factor influencing the SeEvo method’s performance in optimizing DFJSSP. To identify the optimal parameter combination, this study applies the Taguchi orthogonal experimental design. The SeEvo method involves four parameters: population size, maximum number of iterations, mutation rate, and LLM temperature, as detailed in Table~\ref{tab:addlabe0}. The parameter settings are as follows: Population size\(\in \{10, 15, 20\}\), Maximum iterations \(\in \{10, 15, 20\}\), Mutation rate\(\in \{0.5, 0.7, 0.9\}\), and LLM temperature\(\in \{0, 0.5, 1\}\).

\begin{figure}[htbp]
	\centering
	\includegraphics[width=\linewidth]{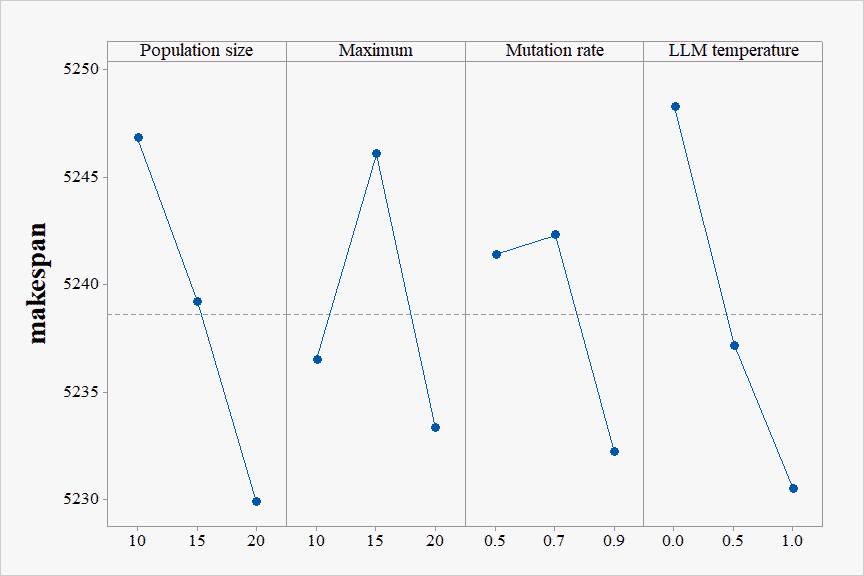}			
	\caption{Main effect plot for SeEvo on four parameters.}
	\label{tiankou}
\end{figure}
To achieve this, an \(L_9\) orthogonal array is generated, and each parameter combination is independently evaluated three times across four training cases (two static cases and two dynamic cases). As shown in Fig.\ref{tiankou}, the experimental analysis of the average makespan, conducted using Minitab 18, reveals that the optimal parameter combination is Population size = 20, Maximum iterations = 20, Mutation rate = 0.9, and LLM temperature = 1.0. It is worth noting that the selected values for population size and maximum number of iterations are relatively small. In our Taguchi experiment, we intentionally constrain parameters such as population size and maximum iterations to strike a balance between performance improvements and the computational costs associated with LLM API calls. The primary goal is to optimize HDRs and prompts to efficiently generate effective dynamic scheduling solutions for the test set. Additionally, achieving high-quality dynamic scheduling results with low computational costs remains a promising direction for future research.

\subsection{Generalization Performance on Public Benchmark Datasets for Makespan Optimization}
In this section, we evaluate the performance of the SeEvo method using 160 benchmark cases from the TA and DMU datasets, covering 8 distinct problem sizes. The TA test set sizes range from \(15 \times 15\) to \(100 \times 20\), while the DMU test set sizes range from \(20 \times 15\) to \(50 \times 20\).

\textbf{Performance on the DMU Benchmark:} This section compares seven well-known HDRs, including Random selection, shortest processing time (SPT), shortest total processing time (STPT), most process sequence remaining (MPSR), most work remaining (MWKR), minimum ratio of flow due date to most work remaining (FDD/MWKR), most operations remaining (MOPNR), as well as MTGP, GEP, DRL-Liu \cite{liuDynam2024}, DRL-Zhang \cite{zhangLearn2020c}, and SeEvo. The results in Table \ref{tab:addlabel} compare SeEvo with HDRs and DRL-Liu methods, with benchmark data sourced from \cite{liuDynam2024}. We select this specific subset to ensure fair comparison with these DRL methods, as they use these cases to validate generalization performance. SeEvo outperforms other methods in 14 out of 16 test cases, ranking second in the remaining 2 cases with minimal differences (only 1 unit behind in DMU29). These findings demonstrate SeEvo's strong generalization capabilities across JSSP cases of varying sizes. Table \ref{tab:addlabeladd} summarizes the experimental results across all DMU benchmark datasets, with each value representing the arithmetic mean makespan for cases of the same size. The results consistently show that SeEvo outperforms all alternatives, followed by GP and GEP, which surpass traditional HDRs and the DRL-Zhang \cite{zhangLearn2020c} approach.

\begin{table*}[htbp]
  \centering
  \caption{Experimental Results on DMU Benchmark, where the ``UB'' column indicates the best-known solution}
    \resizebox{0.96\textwidth}{!}{%
    \scriptsize
    \setlength{\tabcolsep}{2pt}%
    \renewcommand{\arraystretch}{0.92}%
    \begin{tabular}{@{}l*{12}{c}@{}}
    \toprule
    Cases & Size & Random & SPT & STPT & MPSR & DRL-Liu & GP & GEP & DeepSeek-V3 & gpt4.1-mini & Qwen-Turbo & UB \\
    \midrule
    DMU03 & \(20 \times 15\) & 3827  & 3630  & 4232  & 3435  & 3303  & 3568  & 3570  & 3383  & \textbf{3238} & 3370  & 2731 \\
    DMU04 & \(20 \times 15\) & 3889  & 3541  & 4642  & 3355  & 3321  & 3381  & 3366  & 3235  & \textbf{3212} & 3235  & 2669 \\
    DMU08 & \(20 \times 20\) & 4228  & 4714  & 4459  & 3999  & 4098  & 3802  & 4023  & \textbf{3728} & \textbf{3728} & \textbf{3728} & 3188 \\
    DMU09 & \(20 \times 20\) & 4094  & 4283  & 4690  & 3869  & \textbf{3753} & 3996  & 4325  & 3828  & 3828  & 3857  & 3092  \\
    DMU13 & \(30 \times 15\) & 5451  & 4813  & 5207  & 4759  & 4708  & \textbf{4647} & 4703  & 4658  & \textbf{4647} & 4709  & 3681  \\
    DMU14 & \(30 \times 15\) & 5306  & 4583  & 4811  & 4238  & 4124  & 4223  & 4074  & 3980  & 3980  & \textbf{3907} & 3394  \\
    DMU18 & \(30 \times 20\) & 5326  & 6231  & 5480  & 5003  & 4800  & 4688  & 4989  & \textbf{4618} & 4661  & 4724  & 3844  \\
    DMU19 & \(30 \times 20\) & 5174  & 5126  & 5203  & 4930  & 4837  & 5230  & 5081  & 4774  & 4836  & \textbf{4715} & 3764  \\
    DMU23 & \(40 \times 15\) & 5948  & 6250  & 6521  & 5383  & 5240  & 5518  & 5427  & \textbf{5151} & 5220  & \textbf{5151} & 4668  \\
    DMU24 & \(40 \times 15\) & 6078  & 5503  & 6595  & 5358  & 5319  & 5664  & 5256  & 5340  & \textbf{5254} & 5316  & 4648  \\
    DMU28 & \(40 \times 20\) & 6737  & 6558  & 7697  & 5927  & 5948  & 6294  & 6142  & 5838  & \textbf{5812} & 5829  & 4692  \\
    DMU29 & \(40 \times 20\) & 6602  & 6565  & 7690  & 6107  & \textbf{5824} & 6243  & 6146  & 5936  & 5825  & 5866  & 4691  \\
    DMU33 & \(50 \times 15\) & 6890  & 7361  & 7631  & 6282  & 6458  & 6140  & 6311  & 6014  & 6029  & \textbf{6009} & 5728  \\
    DMU34 & \(50 \times 15\) & 7523  & 7026  & 7740  & 6359  & 6284  & 6223  & 6364  & 6127  & \textbf{6098} & 6148  & 5385  \\
    DMU38 & \(50 \times 20\) & 7685  & 7954  & 8555  & 7604  & 7275  & 7586  & 7380  & \textbf{7066} & 7197  & 7168  & 5713  \\
    DMU39 & \(50 \times 20\) & 8097  & 7592  & 8908  & 6953  & 6776  & 6754  & 6834  & 6693  & \textbf{6591} & 6604  & 5747 \\
    \multicolumn{2}{l}{Mean} & 5803.44  & 5733.13  & 6253.81  & 5222.56  & 5129  & 5247.31  & 5249.44  & 5023.06  & \textbf{5009.75} & 5021.00  & 4227.19  \\
    \bottomrule
    \end{tabular}%
    }%
  \label{tab:addlabel}
\end{table*}

\begin{table*}[htbp]
  \centering
  \caption{Results on DMU Benchmark based on the arithmetic mean for each problem size, where the ``UB'' column indicates the best-known solution}
    \resizebox{0.96\textwidth}{!}{%
    \scriptsize
    \setlength{\tabcolsep}{2pt}%
    \renewcommand{\arraystretch}{0.92}%
    \begin{tabular}{@{}l*{10}{c}@{}}
    \toprule
    Cases  & MWKR  & FDD/WKR & MOPNR & DRL-Zhang   & GP    & GEP   & DeepSeek-V3 & gpt4.1-mini & Qwen-Turbo & UB \\
    \midrule
    DMU20$\times$15 & 4909.9  & 4666.3  & 4513.2  & 4215.3  & 4171.6  & 4064.9  & \textbf{3868.6} & 3870.4  & 3893.4  & 3023.8  \\
    DMU20$\times$20 & 5489.0  & 5298.2  & 5052.3  & 4804.5  & 4549.9  & 4560.9  & 4339.7  & \textbf{4330.1} & 4338.2  & 3472.6  \\
    DMU30$\times$15 & 6252.9  & 6016.5  & 5742.8  & 5557.9  & 5381.7  & 5343.7  & 5210.2  & \textbf{5185.7} & 5188.2  & 3879.0  \\
    DMU30$\times$20 & 6925.0  & 6827.3  & 6491.9  & 5967.4  & 5833.8  & 5875.6  & \textbf{5604.1} & 5624.3  & 5610.9  & 4248.4  \\
    DMU40$\times$15 & 7484.2  & 7420.0  & 7105.5  & 6663.9  & 6704.0  & 6607.2  & 6400.0  & \textbf{6374.9} & 6385.0  & 4871.2  \\
    DMU40$\times$20 & 8460.9  & 8210.9  & 7870.7  & 7375.8  & 7392.0  & 7368.8  & 7070.1  & \textbf{7042.8} & 7050.1  & 5240.9  \\
    DMU50$\times$15 & 8906.0  & 9150.2  & 8436.5  & 8179.4  & 8190.2  & 8119.7  & \textbf{7854.0} & 7863.5  & 7893.9  & 5950.6  \\
    DMU50$\times$20 & 9807.0  & 9899.6  & 9408.0  & 8751.6  & 8775.1  & 8819.2  & 8496.0  & \textbf{8463.4} & 8473.3  & 6227.3  \\
    \bottomrule
    \end{tabular}%
    }%
  \label{tab:addlabeladd}%
\end{table*}%

\begin{table*}[htbp]
  \centering
  \caption{Experimental Results on TA Benchmark, where the ``UB'' column indicates the best-known solution}
    \resizebox{0.96\textwidth}{!}{%
    \scriptsize
    \setlength{\tabcolsep}{2pt}%
    \renewcommand{\arraystretch}{0.92}%
    \begin{tabular}{@{}l*{11}{c}@{}}
    \toprule
    Cases & Size & SPT/TWKR & DRL-Chen & DRL-Zhang & DRL-Liu & GP & GEP & DeepSeek-V3 & gpt4.1-mini & Qwen-Turbo  & UB \\
    \midrule
    TA01  & \(15 \times 15\) & 1664  & 1711  & 1433  & 1492  & 1539  & 1442  & \textbf{1427} & \textbf{1427} & \textbf{1427} & 1231 \\
    TA02  & \(15 \times 15\) & 1538  & 1639  & 1544  & \textbf{1425} & 1625  & 1458  & 1432  & 1437  & 1460  & 1244 \\
    TA11  & \(20 \times 15\) & 1886  & 1833  & 1794  & 1752  & 1866  & 1819  & 1694  & \textbf{1671} & \textbf{1671} & 1357 \\
    TA12  & \(20 \times 15\) & 1969  & 1765  & 1805  & 1692  & \textbf{1601} & 1690  & 1616  & 1614  & 1637  & 1367 \\
    TA21  & \(20 \times 20\) & 2206  & 2145  & 2252  & 2097  & 2093  & 2125  & 1992  & 1986  & \textbf{1977} & 1642 \\
    TA22  & \(20 \times 20\) & 2111  & 2015  & 2102  & 1924  & 2006  & 2010  & \textbf{1915} & 1954  & 1937  & 1600 \\
    TA31  & \(30 \times 15\) & 2435  & 2382  & 2565  & 2277  & 2274  & 2232  & 2210  & 2201  & \textbf{2156} & 1764 \\
    TA32  & \(30 \times 15\) & 2512  & 2458  & 2388  & \textbf{2203} & 2388  & 2399  & 2245  & 2279  & 2279  & 1784 \\
    TA41  & \(30 \times 20\) & 2898  & 2541  & 2667  & 2698  & 2578  & 2727  & 2504  & \textbf{2496} & 2543  & 2006 \\
    TA42  & \(30 \times 20\) & 2813  & 2762  & 2664  & 2623  & 2536  & 2466  & \textbf{2363} & 2419  & \textbf{2363} & 1939 \\
    TA51  & \(50 \times 15\) & 3768  & 3762  & 3599  & 3608  & 3472  & 3509  & \textbf{3378} & 3412  & 3426  & 2760 \\
    TA52  & \(50 \times 15\) & 3588  & 3511  & 3341  & 3524  & 3409  & 3420  & 3304  & \textbf{3245} & 3300  & 2756 \\
    TA61  & \(50 \times 20\) & 3752  & 3633  & 3654  & 3548  & 3668  & 3709  & 3489  & \textbf{3477} & 3510  & 2868 \\
    TA62  & \(50 \times 20\) & 3925  & 3712  & 3722  & 3557  & 3577  & 3707  & \textbf{3446} & 3474  & \textbf{3446} & 2869 \\
    TA71  & \(100 \times 20\) & 6705  & 6321  & 6452  & 6289  & 6247  & 6204  & 6071  & \textbf{6059} & 6071  & 5464 \\
    TA72  & \(100 \times 20\) & 6351  & 6232  & 5695  & 6002  & 5790  & 5660  & 5604  & \textbf{5579} & 5591  & 5181 \\
    \multicolumn{2}{l}{Mean} & 3132.56  & 3026.38  & 2979.81  & 2919.44  & 2916.81  & 2911.06  & \textbf{2793.13} & 2795.63  & 2799.63  & 2364.50  \\
    \bottomrule
    \end{tabular}%
    }%
  \label{tab:addlabe2}%
\end{table*}

\begin{table*}[htbp]
  \centering
  \caption{Results on TA Benchmark based on the arithmetic mean for each problem size, where the ``UB'' column indicates the best-known solution}
    \resizebox{0.96\textwidth}{!}{%
    \scriptsize
    \setlength{\tabcolsep}{2pt}%
    \renewcommand{\arraystretch}{0.92}%
    \begin{tabular}{@{}l*{10}{c}@{}}
    \toprule
    Cases  & MWKR  & FDD/WKR & MOPNR & DRL-Zhang   & GP    & GEP   & DeepSeek-V3 & gpt4.1-mini & Qwen-Turbo & UB \\
    \midrule
    TA15$\times$15 & 1927.5  & 1808.6  & 1782.3  & 1547.4  & 1543.2  & 1528.7  & 1448.1  & \textbf{1440.8} & 1446.7  & 1228.9 \\
    TA20$\times$15 & 2190.7  & 2054.0  & 2015.8  & 1774.7  & 1763.2  & 1774.6  & 1678.1  & \textbf{1657.5} & 1666.3  & 1364.9 \\
    TA20$\times$20 & 2518.6  & 2387.2  & 2309.9  & 2128.1  & 2075.2  & 2083.5  & 1951.2  & 1947.8  & \textbf{1943.3} & 1617.5 \\
    TA30$\times$15 & 2728.0  & 2590.8  & 2601.3  & 2378.8  & 2342.8  & 2350.0  & 2207.3  & 2209.0  & \textbf{2192.4} & 1790.2 \\
    TA30$\times$20 & 3193.3  & 3045.0  & 2888.1  & 2603.9  & 2611.1  & 2588.4  & 2450.2  & 2448.6  & \textbf{2447.1} & 1948.5 \\
    TA50$\times$15 & 3907.8  & 3736.3  & 3608.0  & 3393.8  & 3347.2  & 3386.3  & 3217.6  & \textbf{3214.0} & 3223.8  & 2773.8 \\
    TA50$\times$20 & 4375.1  & 4022.1  & 3920.0  & 3593.9  & 3532.2  & 3574.0  & \textbf{3348.7} & 3362.9  & 3350.9  & 2843.9 \\
    TA100$\times$20 & 7128.8  & 6620.7  & 6452.3  & 6097.6  & 6024.3  & 6037.8  & 5901.3  & \textbf{5881.3} & 5888.7  & 5365.7 \\
    \bottomrule
    \end{tabular}%
    }%
  \label{tab:addlabe2add}%
\end{table*}

\textbf{Performance on the TA Benchmark:} SeEvo's performance is also evaluated on the TA benchmark, comparing with three recent DRL-based approaches (DRL-Chen \cite{chendeep2023}, DRL-Zhang \cite{zhangLearn2020c}, DRL-Liu \cite{liuDynam2024}) and four HDRs: SPT/TWKR, MWKR, FDD/MWKR, and MOPNR, along with GP and GEP. The results in Table \ref{tab:addlabe2} source HDRs data from \cite{chendeep2023} and DRL data from \cite{liuDynam2024}. SeEvo outperforms other approaches in 13 out of 16 test cases, ranking second in the remaining 3 cases. Table \ref{tab:addlabe2add} demonstrates SeEvo's generalization ability across all TA datasets, with values representing the arithmetic mean makespan for cases of the same size.

Across both DMU and TA benchmarks, SeEvo consistently demonstrates superior performance, showcasing its ability to generate high-quality scheduling solutions even when problem sizes and processing times differ from training data. Among the three LLMs, gpt-4.1-mini performs best, followed by DeepSeek-V3 and Qwen-Turbo. SeEvo's strong performance stems from effective prompt engineering and domain knowledge gained from iterative training, enabling efficient generation of high-quality scheduling decisions. On test sets with 100 jobs and 20 machines, Qwen-Turbo and gpt-4.1-mini generate solutions in 40 and 42 seconds respectively, while DeepSeek-V3 requires 122 seconds. Despite the longer inference time, DeepSeek-V3 offers substantially lower API costs (approximately one-third of gpt-4.1-mini), making it practical for cost-sensitive applications where response time is not critical.

However, as shown in Tables \ref{tab:addlabel}, \ref{tab:addlabeladd}, \ref{tab:addlabe2}, and \ref{tab:addlabe2add}, a gap remains between our method's results and optimal solutions on these static cases. Future work could deploy local LLMs for low-cost iterative collection of HDRs and prompts, combined with vector databases to match similar training instances, further enhancing optimization capabilities. This approach has been validated in a \textit{Nature} publication \cite{romera-paredesMathe2024}, which demonstrated optimal solutions for online bin packing problems.

\subsection{Performance Evaluation in Dynamic Cases for Makespan Optimization}
\begin{figure*}[htbp]
	\centering
	\includegraphics[width=\linewidth]{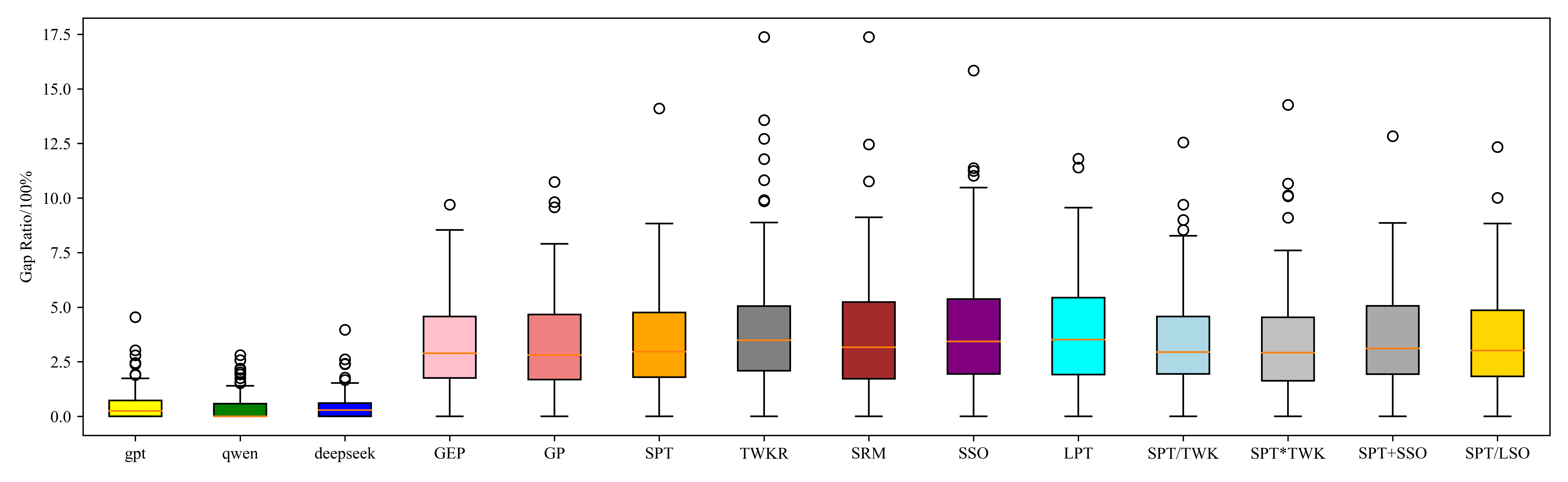}			
	\caption{Gap ratio of different scheduling methods in DFJSSP scenarios for makespan optimization.}
	\label{fig3}
\end{figure*}

In this section, the performance of the SeEvo method is evaluated in a dynamic environment characterized by fuzzy processing times, randomly arriving orders, as well as machine breakdowns and repairs. The evaluation is conducted on 100 test case to comprehensively assess the method's robustness and effectiveness. The results, presented in Fig. \ref{fig3}, compare SeEvo with 9 common HDRs, including SPT, TWKR, shortest remaining machining time (SRM), shortest subsequent operation (SSO), LPT, LPT/TWK, SPT$\times$TWK, SPT+SSO, SPT/LSO, GP, and GEP. To normalize the performance across different problem instances, we calculate the relative percentage gap (Gap Ratio) with the following formula:

\begin{equation}
\text{Gap Ratio} \left(\%\right) = \frac{Obj_{i,j} - Obj_{i,best}}{Obj_{i,best}} \times 100%
\end{equation}
where $Obj_{i,j}$ represents the objective value obtained by algorithm $j$ on problem case $i$, and $Obj_{i,best}$ is the minimum objective value among all 14 compared methods for that case. The performance of all methods, evaluated using this metric, is presented in Fig. \ref{fig3}.

As shown in Fig.\ref{fig3}, SeEvo consistently demonstrates significant advantages under dynamic conditions, achieving substantial performance improvements in most cases. The method exhibits the smallest relative gap, with the maximum gap ratio not exceeding 5\%, and in most cases, remaining below 1\%. These results indicate that SeEvo can effectively generate good scheduling solutions across various DFJSSP scenarios, even in the presence of fuzzy processing times, random order arrivals, and machine breakdowns. Overall, SeEvo showcases a strong ability to handle dynamic environments and uncertainties, closely resembling real-world manufacturing conditions.

The excellent performance of SeEvo can be attributed to several factors. Firstly, the SeEvo framework integrates extensive domain knowledge, enabling the generation of generalized HDRs and prompts. By training on 20 randomly sized order batches, these HDRs enhance the model's ability to generalize to new scenarios, allowing effective decision-making with a single iteration. Secondly, SeEvo improves both exploration and exploitation through individual co-evolution reflection, individual self-evolution reflection, and collective evolution reflection, ensuring high-quality HDRs are generated.

\subsection{Performance Evaluation in Dynamic Cases for Tradiness Optimization}
\begin{figure*}[htbp]
	\centering
	\includegraphics[width=\linewidth]{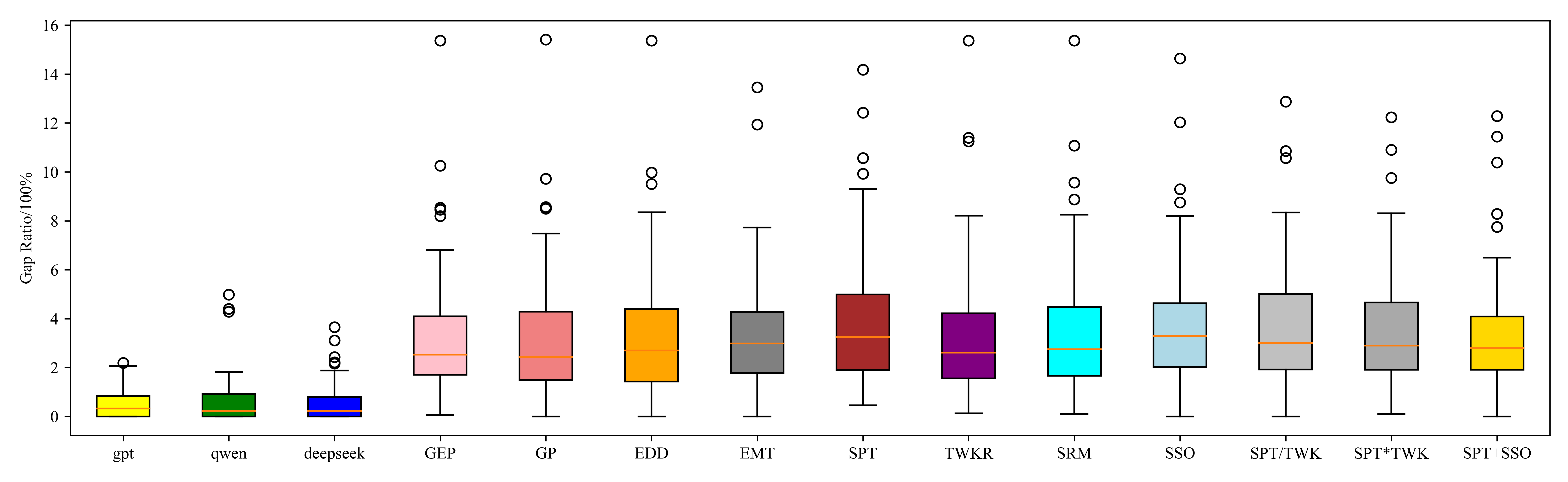}			
	\caption{Gap Ratio of different scheduling methods in DFJSSP scenarios for tardiness optimization.}
	\label{fig3_add1}
\end{figure*}

This section evaluates the tardiness optimization performance of the SeEvo method in dynamic environments, with fuzzy processing times, random job arrivals, and machine breakdowns and repairs. The evaluation is conducted on 100 test cases to comprehensively assess the method's performance. The due date of each job is defined as the sum of its operation processing times. For newly arriving orders, the due date is calculated as the arrival time plus the total processing time of all operations. The comparative tardiness results are shown in Fig.\ref{fig3_add1}, where SeEvo is compared with 9 common HDRs, including earliest due date (EDD), estimated maximum tardiness (EMT), SPT, TWKR, SRM, SSO, LPT/TWK, SPT$\times$TWK, SPT+SSO, GP, and GEP. Fig.\ref{fig3_add1} presents the gap ratio between each method and the optimal results, where the optimal result is the best result among the 14 methods for each case, not the true optimal solution.

As shown in Fig.\ref{fig3_add1}, SeEvo consistently demonstrates significant advantages under dynamic conditions, achieving substantial improvements in performance in most scenarios. The proposed method achieves the smallest relative gap, with the maximum gap ratio not exceeding 6\%, and in most cases, remaining below 1\%. These results indicate that SeEvo can generate good scheduling results, even with random job arrivals and machine breakdowns,  further validating its generalization capability in solving DFJSSP cases. Among the evaluated methods, SeEvo slightly outperforms DeepSeek-V3 and Qwen-Turbo, and is superior to GP, GEP, and other HDRs.

In our experimental setup, we intentionally set shorter due dates to create clearer distinctions in tardiness levels across methods. Longer due dates could result in many cases with zero tardiness, limiting the ability to differentiate method performance. This choice ensures measurable performance differences across all test cases. However, we acknowledge that real-world manufacturing environments typically feature more dynamic due dates, influenced by shop floor conditions, production constraints, and customer requirements.

\subsection{Ablation Study}
The core innovation of the proposed SeEvo method lies in the introduction of individual self-evolution. To evaluate its impact, an ablation study is conducted by removing this process, resulting in a modified method referred to as ReEvo. The exploration and exploitation capabilities of both SeEvo and ReEvo are assessed through 50 iterations on static training datasets, followed by testing their generalization performance.

\begin{figure*}[htbp]
	\centering
	\includegraphics[width=\linewidth]{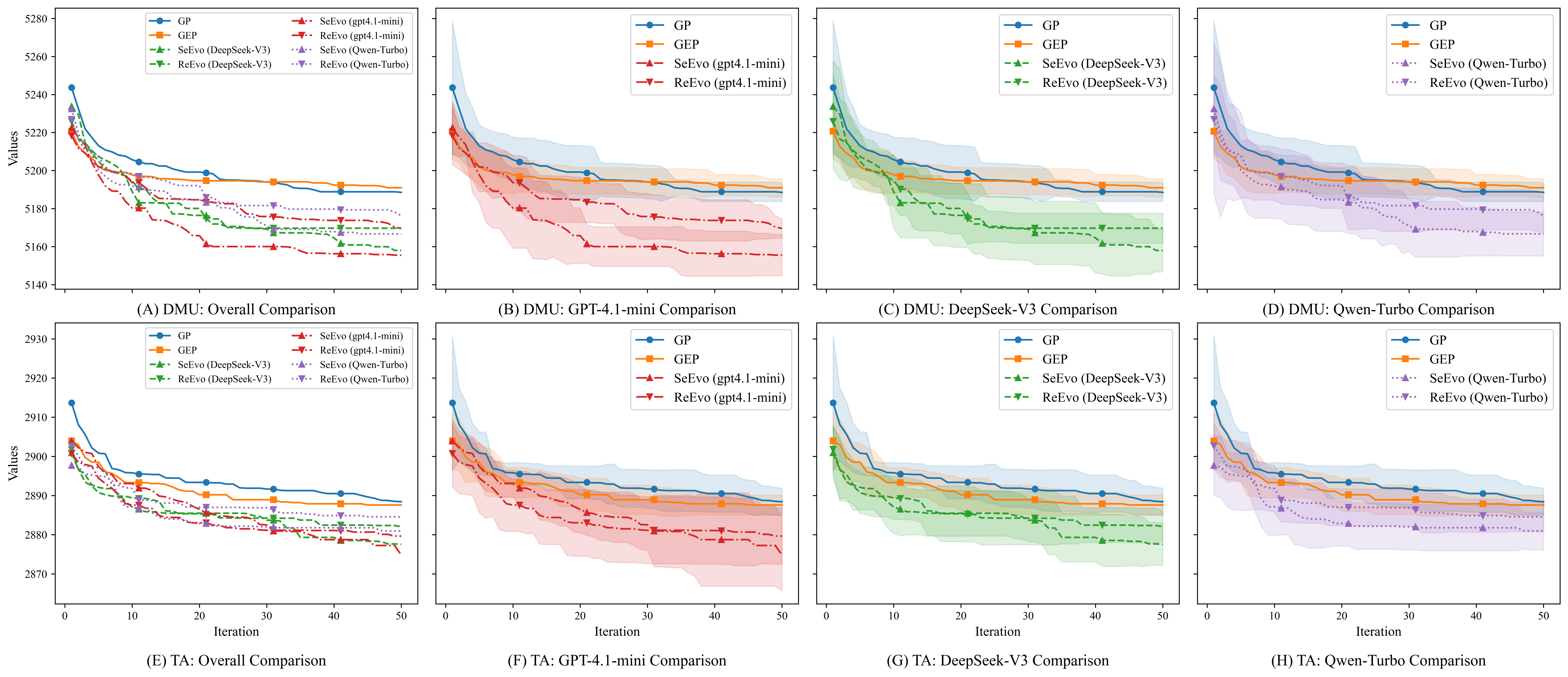}			
	\caption{Convergence Curves of Different Methods on Two Training Case Groups.}
	\label{fig4}
\end{figure*}

Training is conducted on the datasets presented in Tables \ref{tab:addlabel} and \ref{tab:addlabe2}. Each method is independently executed 10 times, and Fig. \ref{fig4} illustrates the convergence curves with mean values and standard deviations (shaded areas) to assess the stability of the results. The x-axis represents the number of validation iterations, while the y-axis denotes the average makespan across the datasets. As shown in Fig. \ref{fig4}, all LLM-based methods significantly outperform the traditional GP and GEP baselines, demonstrating that guided exploration is substantially more effective than random search. More importantly, SeEvo variants consistently outperform their ReEvo counterparts in the mid-to-late iteration stages across both datasets. The shaded areas representing standard deviations reveal that this performance gap is more pronounced on the complex DMU dataset (Subfigures B, C, and D), where the convergence curves of SeEvo and ReEvo exhibit limited overlap after 15-20 iterations. In contrast, on the simpler TA task (Subfigures F, G, and H), the higher degree of overlap between methods suggests that the advantages of self-evolution become particularly evident when addressing complex scheduling problems.

The generalization performance of ReEvo is further evaluated on the TA test sets for two problem sizes: 30 jobs with 20 machines and 50 jobs with 20 machines. Each problem size included 10 cases, with the optimization objective being the minimization of completion time. The results in Table \ref{tab:addlabe3} show that SeEvo outperforms ReEvo as well as two end-to-end DRL methods \cite{liuDynam2024}, demonstrating the critical role of individual self-evolution in enhancing exploration. The superior performance of SeEvo over DRL methods in generalization scenarios can be attributed to its self-evolutionary mechanism. Unlike DRL approaches that rely on fixed neural network policies trained on specific data distributions, SeEvo continuously evolves its HDR library through iterative refinement guided by LLM reasoning. This allows SeEvo to leverage both the accumulated diverse rule repository and the LLM's contextual understanding to adaptively generate rules tailored to unseen scenarios. The combination of evolutionary diversity and semantic reasoning enables SeEvo to better anticipate and respond to variations in problem scales and processing time distributions.
\begin{table*}[htbp]
  \centering
  \caption{Ablation Study Results}
  \resizebox{0.96\textwidth}{!}{%
  \scriptsize
  \setlength{\tabcolsep}{2pt}%
  \renewcommand{\arraystretch}{0.92}%
  \begin{tabular}{@{}l*{8}{c}@{}}
    \toprule
    Instance  & DRL-GAT & DRL-GCN & SeEvo(DeepSeek) & SeEvo(gpt4.1) & SeEvo(Qwen) & ReEvo(DeepSeek) & ReEvo(gpt4.1) &  ReEvo(Qwen)\\
    \midrule
    30$\times$20 & 2628  & 2740  & 2450.2 & 2448.6 & \textbf{2447.1} & 2456.8 & 2452.1 & 2461.5  \\
    50$\times$20 & 3547  & 3703  & \textbf{3348.7} & 3362.9 & 3350.9 & 3365.2 & 3367.5 & 3378.8 \\
    \bottomrule
  \end{tabular}%
  }%
  \label{tab:addlabe3}%
\end{table*}

\begin{figure}[htbp]
	\centering
	\includegraphics[width=\linewidth]{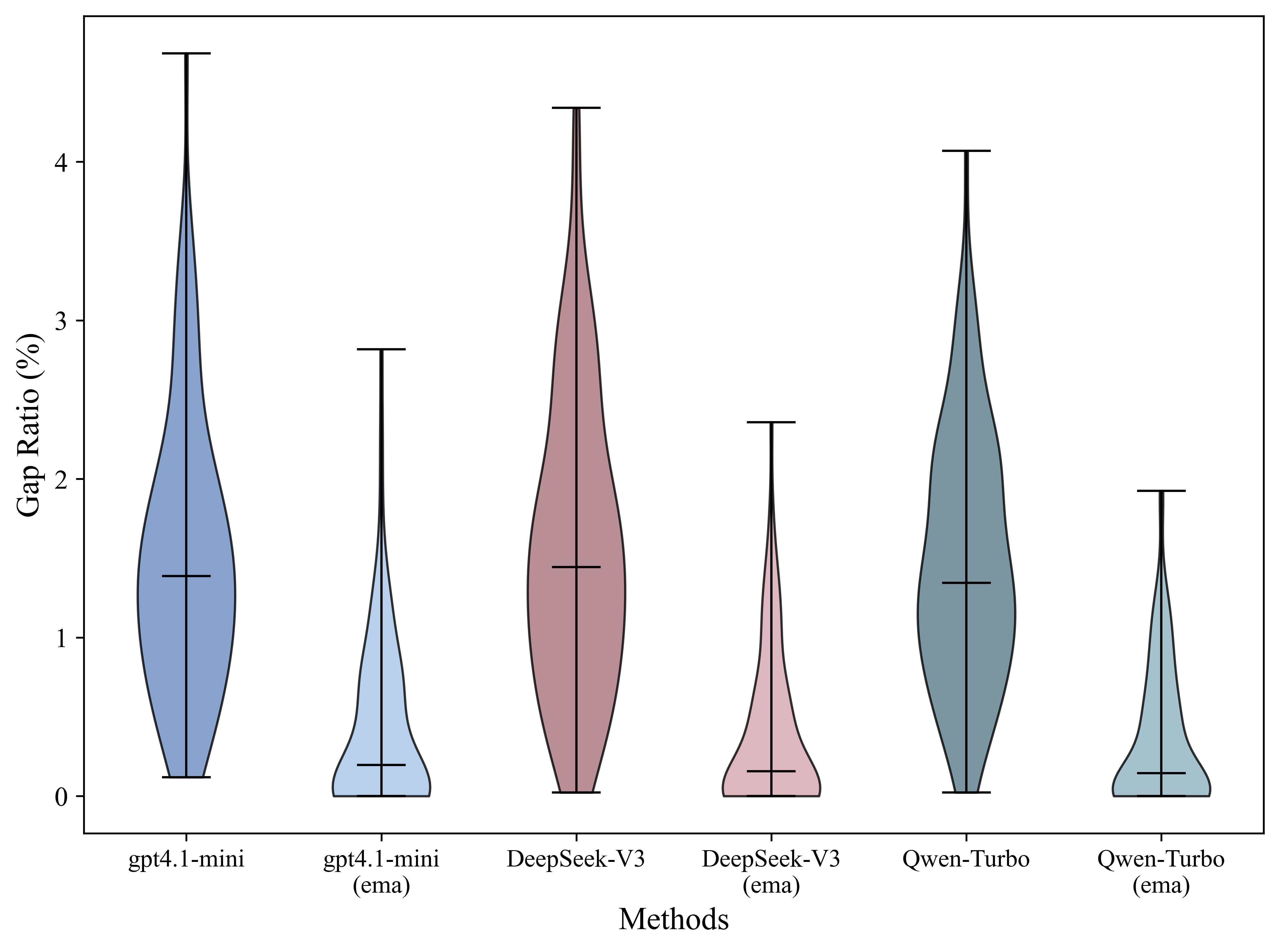}			
	\caption{EMA feature ablation: Gap ratio distribution and stability.}
	\label{fig4_add}
\end{figure}

\begin{figure}[htbp]
	\centering
	\includegraphics[width=\linewidth]{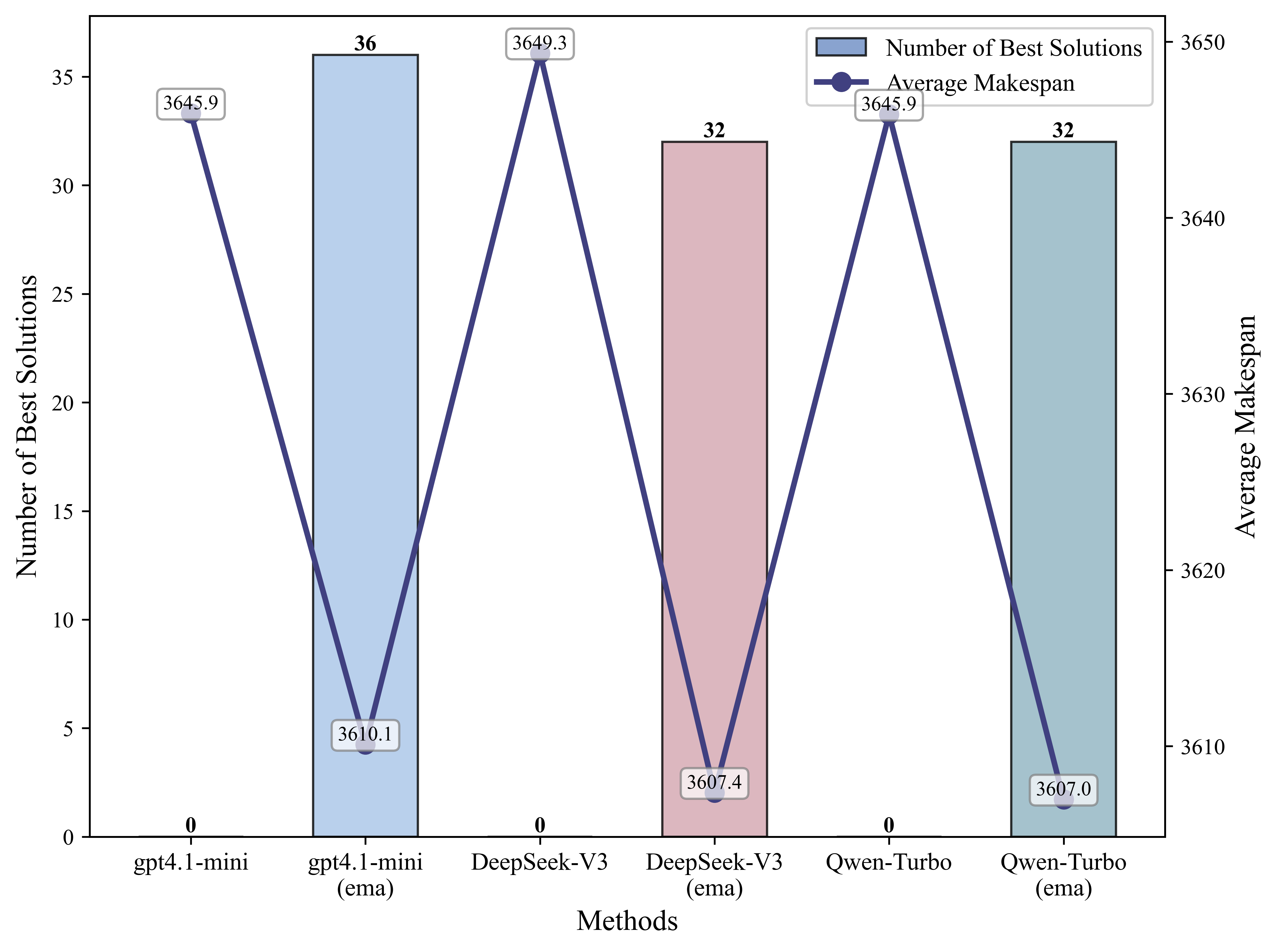}			
	\caption{EMA feature ablation: best solutions quality and average Makespan.}
	\label{fig4_add1}
\end{figure}

A second ablation study is conducted to validate the ema feature, our mechanism for addressing processing time randomness. On 100 randomly generated dynamic instances, we compare the performance of heuristics generated with and without this feature. The results are detailed in Fig. \ref{fig4_add} and Fig. \ref{fig4_add1}. As shown in the violin plots (Fig. \ref{fig4_add}), policies incorporating ema achieved both a significantly lower median Gap (better performance) and a narrower distribution, demonstrating enhanced robustness. This is corroborated by Fig. \ref{fig4_add1}, where the SeEvo with ema approach secured a far greater number of best solutions and a consistently lower average makespan. These findings empirically confirm that the ema feature is critical for generating robust heuristics that effectively mitigate uncertainty.

\begin{figure}[htbp]
	\centering
	\includegraphics[width=\linewidth]{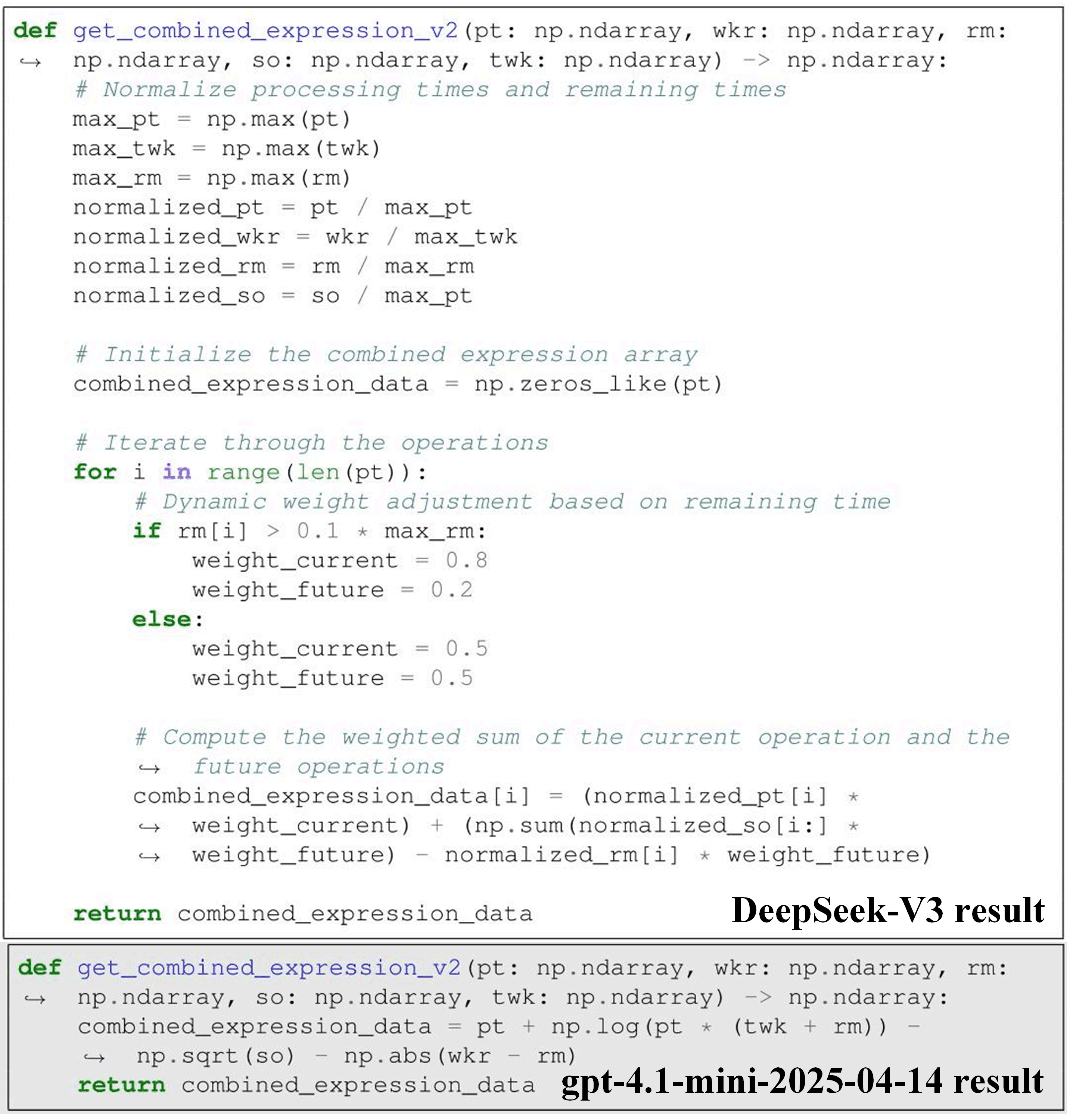}			
	\caption{HDRs generated by Two API Models on TA72 Case.}
	\label{fig5}
\end{figure}

\begin{figure}[htbp]
	\centering
	\includegraphics[width=\linewidth]{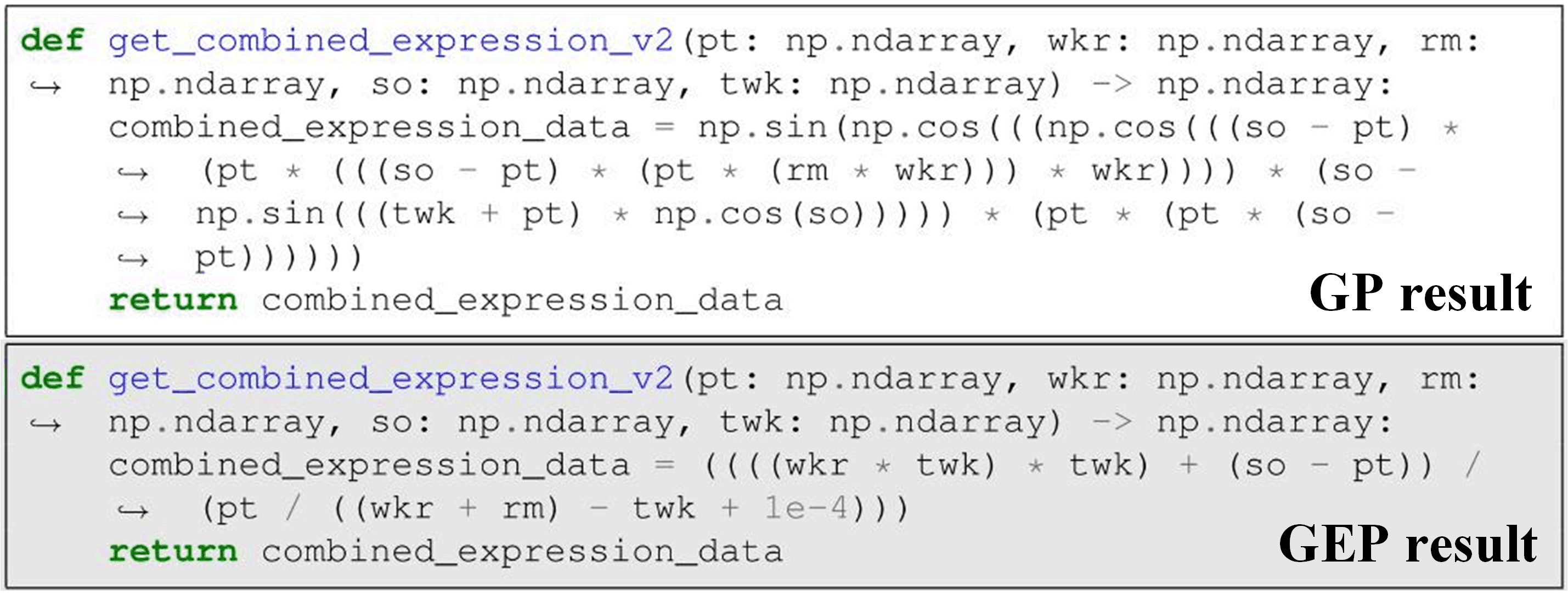}			
	\caption{HDRs generated by GP and GEP on TA72 Case.}
	\label{fig5_add}
\end{figure}

Fig.~\ref{fig5} shows the HDRs generated by two models. The lighter upper part represents the HDR from DeepSeek-V3, while the darker part corresponds to the HDR from GPT-4.1-mini-2025-04-14. Fig.~\ref{fig5_add} shows HDRs generated by GP/GEP. Human-designed HDRs, such as SPT, LPT, and EDD, are tailored for specific scheduling problems but lack scalability and flexibility. In terms of complexity, most generated rules (e.g., from GPT-4.1-mini, GP, and GEP) exhibit $O(n)$ time complexity. However, some rules, like those from DeepSeek-V3, can be less efficient $(O(n^2))$ due to iterative structures. While the GP-generated rule has $O(n)$ complexity, its extreme nesting leads to high computational overhead and limited interpretability. In contrast, SeEvo-generated HDRs strike a balance, being both computationally efficient and more logical. Unlike DRL methods \cite{liuDynam2024}, SeEvo significantly reduces the reliance on high-performance GPUs. Previous DRL methods in experiments used an Intel Core i7 4.0 GHz CPU and a NVIDIA GeForce GTX TITAN X GPU \cite{liuDynam2024}, whereas SeEvo uses no reliance on high-end GPUs.  Although external API usage incurs some cost, SeEvo primarily depends on the enhanced analytical capabilities of LLMs. The extensive parameter space of LLMs provides greater creativity and problem-solving potential. By utilizing localized LLMs, we can gain significant creative advantages, though high-performance GPUs, such as the NVIDIA GeForce RTX 4090D, are still necessary for fully exploiting these capabilities.

Nevertheless, it is important to note that the superiority of a particular scheduling method is highly context-dependent and should be evaluated based on the specific characteristics of each case. We believe that no single dynamic scheduling approach currently demonstrates absolute superiority over all others across all scenarios. Different methods may excel under different problem configurations, constraints, and operational conditions. However, LLM-based approaches represent a promising research direction that warrants further exploration by the research community, particularly given their potential for semantic understanding and adaptive reasoning in complex scheduling environments.

\section{Conclusion and Future work}

To address the challenges of poor generalization and reliance on random search in automatic algorithm design for the DFJSSP, this paper proposes a dynamic evolutionary framework leveraging LLMs. Specifically, the innovative SeEvo method is introduced and comprehensively validated through both static and dynamic fuzzy processing time case studies, with a primary focus on makespan optimization. The experimental results show that SeEvo outperforms commonly used HDRs, GP, GEP, and DRL methods in terms of static generalization performance. Additionally, in dynamic fuzzy processing time experiments, SeEvo exhibits substantial advantages in most cases. This success in handling fuzzy environments is a direct result of our core contribution: a learning framework where the LLM is trained to anticipate the impact of time deviations. By leveraging the $ema$ feature to track historical uncertainty trends, the SeEvo-generated HDRs develop a crucial robustness that is absent in conventional approaches. We also extend our evaluation to include tardiness optimization, where SeEvo consistently shows notable advantages over other methods such as HDRs, GP, and GEP. These findings highlight the potential of LLMs in generating adaptive HDRs, benefiting from their advanced language understanding and generative capabilities to address scheduling under uncertainty.

Despite these promising results, there are limitations. LLMs are used via costly external LLM APIs calls without direct fine-tuning for specific DFJSSP knowledge. Moreover, only a single HDR is employed in dynamic cases, which limits the method's performance, preventing it from achieving the absolute superiority observed in the static cases. Future research will focus on two key areas: (1) We will transition to locally deployed LLMs (e.g., Qwen-7B/8B). This strategic move is crucial not only to reduce computational costs but also to enable the larger-scale experiments and hyperparameter exploration suggested by our findings. We will also fine-tune these models with techniques like direct preference optimization to enhance the domain-specificity of generated HDRs. (2) To improve generalization on static datasets, we will incorporate a SimGNN-based approach to select optimal HDRs by matching the graph-structural similarity of new problems with a case database. (3) Enhancing adaptability to dynamic environments by utilizing a combination of multiple HDRs, thus enabling further optimization of performance in real-time scenarios.

\bibliographystyle{IEEEtran}
\bibliography{refs}

%
%
%
%
%
%
%
%
%
%
\end{document}